\documentclass[letterpaper, 10 pt, conference]{ieeeconf}  
\usepackage{amsmath,amsfonts}
\usepackage{algorithmic}
\usepackage{algorithm}
\usepackage{array}
\usepackage{textcomp}
\usepackage{stfloats}
\usepackage{url}
\usepackage{verbatim}
\usepackage{cite}
\usepackage{caption}
\usepackage{subcaption}
\usepackage{graphicx}
\usepackage[table]{xcolor}
\usepackage{booktabs}

\IEEEoverridecommandlockouts
\begin{document}

\title{\LARGE \bf Safety-Critical Ergodic Exploration in Cluttered Environments via Control Barrier Functions}

\author{Cameron Lerch, Dayi Dong, and Ian Abraham
\thanks{All authors are with the Department of Mechanical Engineering and Materials Science, Yale University, New Haven 06511, USA. Corresponding author {\tt\small cameron.lerch@yale.edu}}%
}

\maketitle
\thispagestyle{empty}
\pagestyle{empty}
\begin{abstract}
    In this paper, we address the problem of safe trajectory planning for autonomous search and exploration in constrained, cluttered environments. 
    Guaranteeing safe (collision-free) trajectories is a challenging problem that has garnered significant due to its importance in the successful utilization of robots in search and exploration tasks.
    This work contributes a method that generates guaranteed safety-critical search trajectories in a cluttered environment. 
    Our approach integrates safety-critical constraints using discrete control barrier functions (DCBFs) with ergodic trajectory optimization to enable safe exploration.
    Ergodic trajectory optimization plans continuous exploratory trajectories that guarantee complete coverage of a space. 
    We demonstrate through simulated and experimental results on a drone that our approach is able to generate trajectories that enable safe and effective exploration. Furthermore, we show the efficacy of our approach for safe exploration using real-world single- and multi- drone platforms. 

\end{abstract}


\section{Introduction} \label{sec:intro}

    In autonomous search and rescue tasks, robots need to plan effective exploratory trajectories while avoiding potential hazards to ensure continued operation.
    Balancing both the effectiveness of search and the safety of the robot then becomes a challenge as the environment becomes more complex and cluttered. 
    As available free space is reduced, the ability of the robot to reason about where to venture next becomes limited and safety becomes a higher priority leading to ineffective search behaviors. 
    To address this problem, we present a safe trajectory planning method for autonomous search in constrained, cluttered environments through integrated development of ergodic trajectory optimization methods~\cite{Mathew_Mezic_2011_Ergodicity, Miller_Murphey_2016_ErgodicExploration, Salman_Ayvali_Choset_2017_ErgodicObstacles} with safety-critical control approaches~\cite{Manjuanth_Nguyen_2021_SafePlanningCBF, Zeng_Zhang_Sreenath_2021_MPC_DiscreteCBF, Sreenath_2021_EnhancingSafetyMPCwithCBF}. 


    Ergodic trajectory optimization methods, often referred to as ergodic search (or exploration)~\cite{Mathew_Mezic_2011_Ergodicity, Miller_Murphey_2016_ErgodicExploration}, have emerged as exploration methods with the guarantee of complete coverage over a space, irrespective of the spatial scale of the space~\cite{Mezic_Scott_Redd_2009_DerivationErgodicityDiffScales}.
    These methods cast the problem of exploration over a space as a continuous trajectory optimization problem using time-averaged distributional representations of trajectories. 
    The optimization leverages spectral methods to synthesize continuous exploration trajectories where the average time spent in a region is proportional to the measure of importance\footnote{Often referred to as a measure of information or an information measure.} assigned to the region. 
    In addition, some recent adaptations of ergodic search methods have shown the ability to avoid obstacles~\cite{Salman_Ayvali_Choset_2017_ErgodicObstacles, Salman_2018_ErgodicConstrained}; however, they do not provide formal guarantees that the robot will remain in a safe set of states. This leads to trajectories that can violate safety conditions and risk collisions with obstacles, themselves, or other robots. 
    Having formal guarantees prevents an imbalance of task priority which would ultimately place the robot at risk, but impede on task performance. 
    Therefore, this work develops an integrated method that ensures both a complete search over a space and that the generated trajectory ensures robot safety.

    \begin{figure}
        \centering
        \includegraphics[width=0.48\textwidth]{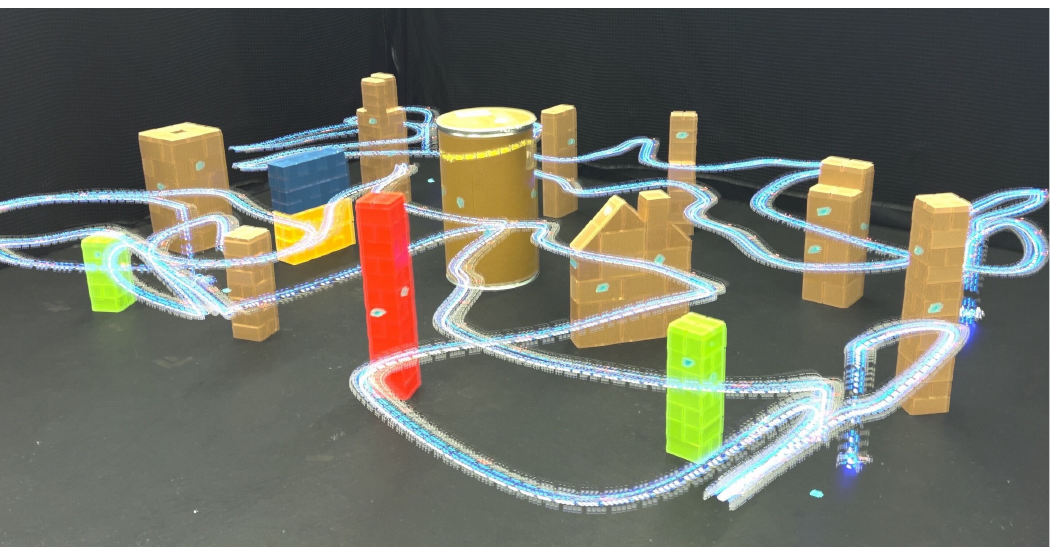}
        \caption{\textbf{Safe, Multi-Robot Ergodic Exploration: } The proposed Safety-Critical Ergodic Trajectory Optimization (SC-ETO) applied to a multi-robot system in a cluttered environment. Our approach generates safe exploratory paths that provide full coverage over a cluttered region while avoiding safely collision with other robots.}
        \label{fig:multi1}
        \vspace{-5mm}
    \end{figure}
    
    Our approach poses safe trajectory optimization for exploration as a constrained ergodic trajectory optimization problem. 
    We leverage discrete-time control barrier functions (DCBFs) as constraints to ensure the safety of the robot along a trajectory and jointly optimize an ergodic trajectory subject to robot motion constraints to enforce effective ergodic coverage.
    We demonstrate in both experiment and simulation that our approach generates safe, ergodic exploratory trajectories in cluttered search environments.
    In addition, we test the robustness of our method in simulation and compare our method with existing approaches. 
    Furthermore, we show that our method is able to generate efficient search trajectories for scenarios with multi-robot exploration (see Fig.~\ref{fig:multi1}).
    
    \noindent Thus, in summary, our contributions are: 
\begin{enumerate}
    \item A method for safety-critical ergodic trajectory optimization (SC-ETO) that integrates discrete-time control barrier functions with ergodic search; and
    \item Demonstrations of our approach on a single- and multi- robot exploration task in a cluttered environment.
\end{enumerate}

    The paper is organized as follows: Section~\ref{sec:background} provides some background on related work.
     Section~\ref{sec:cbf} and Section~\ref{sec:ergodic_search} present preliminary information on safety-critical control via control-barrier functions and ergodic search methods respectively. 
    Section~\ref{sec:methods} derives our proposed method for safety-critical ergodic exploration. 
    Results and conclusions are then presented in Section~\ref{sec:results} and Section~\ref{sec:conclusion}.

\section{Related Work} \label{sec:background}
    
    \noindent
    \textbf{Safety Critical Control: } Planning safe trajectories for search and exploration is a fundamental problem in robotics that involves ensuring the robot remains in a \textit{safe} set of states throughout operation, from an initial configuration to a final (goal) configuration~\cite{Jacopin_Osanlou_2022_PathPlanningReview, SanchesIbanez_2021_PathPlanningReview, Manjuanth_Nguyen_2021_SafePlanningCBF, Sreenath_2021_EnhancingSafetyMPCwithCBF, Yang_Vang_Serlin_Belta_Tron_2019_MotionPlanningCBF}.
    Generally, these problems take the form of obstacle avoidance that guarantee planned trajectories are safe for the robot to navigate during operation. Within the literature, control barrier functions (CBFs) are widely used to enforce these safety-critical constraints on robotic systems \cite{Romdlony_Jayawardhana_2014_UnitingLyapunovandCBF, Manjuanth_Nguyen_2021_SafePlanningCBF}. 
    They have been shown to be an effective way to generate safe trajectories in tight-fitting cluttered environments of polytopes \cite{Thirugnanam_Zeng_Sreenath_2021_ObstacleAvoidance}; and have proven to be useful in a wide variety of robotic systems (e.g., locomotion~\cite{Ames_2021_LocomotionCBF, Ames_Hutter_Grandia_2021_LeggedRobotsCBF, Sreenath_Agrawal_2017_discreteCBF}, automotive~\cite{Ames_Sreenath_Egerstedt_Tabuada_2019_CBFTheoryandApp, Ames_Grizzle_Egerstedt_Tabuada_2017_AutomotiveCBF, Zheng_Ren_Ma_2021_ConstrainedRLwithCBF}, aerial~\cite{Sreenath_2018_DynamicUAVsCBF, Egerstedt_Ames_Wang_2017_QuadrotorsDifferentialFlattness}, and collision avoidance in multi-robot systems~\cite{Egerstedt_Ames_Borrmann_2015_CBFSwarm, Egerstedt_2017_CBFMultiRobot, Egerstedt_Ames_Wang_2017_QuadrotorsDifferentialFlattness}) for both static and dynamic obstacles~\cite{Yang_Vang_Serlin_Belta_Tron_2019_MotionPlanningCBF}. 
    CBFs maintain forward-set invariance, which guarantees that once a robot enters the safe set, it will stay within the safe-set, thereby ensuring its safety \cite{Robey_2020_LearningCBF, Wieland_Allgower_2007_SafteyCBF}.

    Within path planning, several state-of-the-art methods utilize variations of CBFs (Kinodynamic Barrier Functions~\cite{Manjuanth_Nguyen_2021_SafePlanningCBF} and Discrete CBFs~\cite{Thirugnanam_Zeng_Sreenath_2021_ObstacleAvoidance}) as constraints incorporated into sampling-based motion planners (such as RRT) to generate safety-critical trajectories~\cite{Yang_Vang_Serlin_Belta_Tron_2019_MotionPlanningCBF}. 
    However, these methods typically focus on point-to-point planning i.e., getting from some starting configuration to some final configuration without concern for efficiency or exploratory coverage. 
    
    \vspace{2mm}
    \noindent
    \textbf{Ergodic Exploration: }
    Within the context of autonomous exploration, there is a need for algorithms that generate trajectories that are efficient and guarantee effective coverage of an environment. 
    Such methods generate trajectories for robots that spend time exploring areas of interest, while still guaranteeing the robot explores unseen areas.
    Recent methods known as ergodic exploration have been shown to be an effective way to explore a space. 
    Ergodic exploration methods balance exploration of new areas and exploitation of known areas by generating trajectories that spend time in regions of interest proportional to the measure of information in those regions.
    As a result, ergodic exploration methods have demonstrated improved information-gathering behavior compared to prior works~\cite{Eagle_1984_MovingTargetConstrainedSearch, Choset_2001_RoboticCoverageSurvey, Abraham_2020_ErgodicMetricActiveLearning, Abraham_Murphey_Mavrommati_2017_RealTimeCoverageTargetLocalErgodicExplo, Miller_Murphey_2015_OptimalPlanning, Miller_Murphey_2016_ErgodicExploration,  Otte_Patel_2020_MultiAgentCoverage, Silverman_Murphey_2013_OptimalPlanningInfoAcquisition, Kabir_Lee_2020_RecedingHorizonErgoExplo}.
    However, few works consider guaranteed safety within ergodic trajectories~\cite{Salman_Ayvali_Choset_2017_ErgodicObstacles}.
    
    The difficulty lies in generating continuous exploratory trajectories while simultaneously respecting safety constraints.
    Earlier works have used the metric itself to have robots avoid obstacles in the environment~\cite{Mathew_Mezic_2011_Ergodicity,Mezic_2017_ErgodicityBasedCoopMultiagentCoverage} whereas others have used inequality constraints with stochastic optimization to avoid objects~\cite{Salman_Ayvali_Choset_2017_ErgodicObstacles}. 
    However, these methods do not explicitly guarantee the safety of the robot. 
    Rather, inequality constraints that encode distances to obstacles are satisfied by generating trajectories that get arbitrarily close to the boundaries of the object. 
    In scenarios where the robot is unable to track these trajectories or there are modeling inaccuracies, the robot may become unsafe. 
    As a result, real-world implementations of ergodic exploration methods on robots are sparse due to the lack of guaranteed safety~\cite{prabhakar2020ergodic, abraham2018decentralized}. 
    This work presents an integrated approach to jointly plan exploratory ergodic trajectories and provide explicit safety-critical guarantees through control barrier functions and demonstrates the effectiveness of the approach on a real robotic system. 
    
    \begin{figure}
        \centering
        \includegraphics[width=0.48\textwidth]{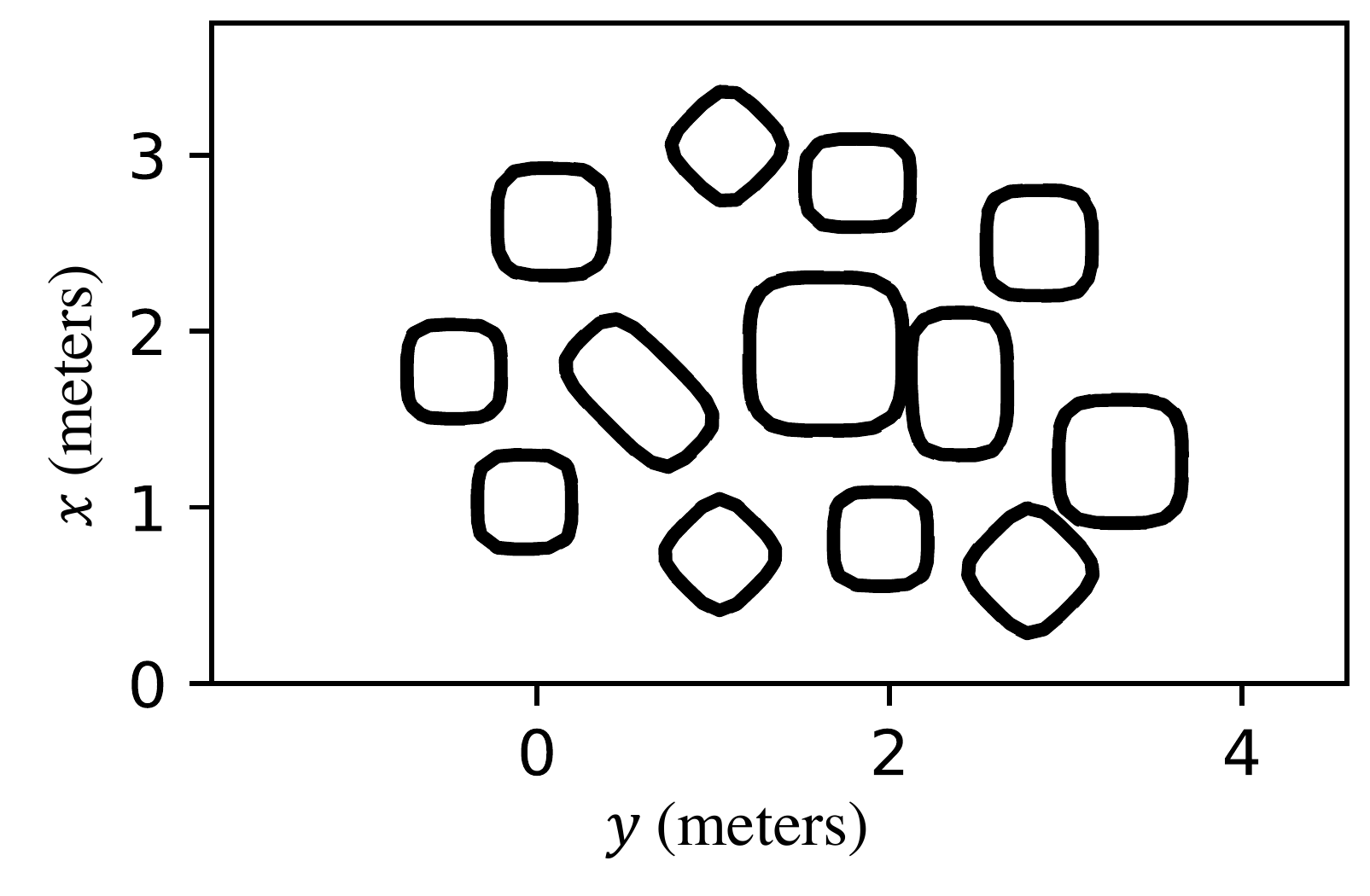}
        \caption{The zero level set of the generated functions $h(x)$ for all of the obstacles in the cluttered environment. Positions of the objects are used for both simulated and real-world experiments.}
        \label{fig:obs_loc}
        \vspace{-5mm}
    \end{figure}
\section{Preliminaries}\label{sec:preliminaries}
    In this section, we present preliminary information on control barrier functions for safety-critical control and ergodic exploration methods for generating exploratory robot trajectories.
    
\subsection{Safety-Critical Control via Control Barrier Functions}\label{sec:cbf}
    
    Let us consider the continuous-time robotic system with states $x \in \mathcal{X} \subset \mathbb{R}^n$ and inputs $u \in \mathcal{U} \subset \mathbb{R}^m$ governed by the differential equation
    \begin{equation}\label{eq:dt-system}
        \dot{x} = f(x, u),
    \end{equation}
    where $f : \mathcal{X} \times \mathcal{U} \to \mathcal{X}$ is the dynamics of the robot and is continuous and differentiable. 
    Next, let us consider the set 
    \begin{equation}
        \mathcal{S} = \{x\in\mathcal{X} \mid h(x) \ge 0 \}
    \end{equation}
    where $h: \mathcal{X} \to \mathbb{R}$ is a continuously differentiable function. 
    The set $\mathcal{S}$ is considered safe if $\forall x \in \mathcal{S}$,
    \begin{equation} \label{eq:cbf_cond}
        \dot{h}(x) = \nabla h(x) \cdot f(x,u) \ge -\gamma(h(x)) \hspace{2mm} \forall u\in \mathcal{U},
    \end{equation}
    and that $\frac{\partial h}{\partial x} \neq 0$ for some $\mathcal{K}_\infty$ function $\gamma$.
    The function $h$ is then known as a control barrier function if Eq.~\eqref{eq:cbf_cond} holds. 
    
    For a discrete-time system, 
    \begin{equation}\label{eq:discrete_time_f}
        x_{t+1} = f(x_t, u_t)
    \end{equation}
    where $f: \mathcal{X}\times \mathcal{U} \to \mathcal{X}$ now evolves the state $x_t$ in time to $x_{t+1}$. The continuous-time CBF expression in Eq.~\eqref{eq:cbf_cond} has been shown to have a discrete-time analog, i.e., a discrete-time control barrier function (DCBF)~\cite{Sreenath_Agrawal_2017_discreteCBF, Zeng_Zhang_Sreenath_2021_MPC_DiscreteCBF}
    \begin{equation}\label{eq:dcbf-def}
        \Delta h(x_t, u_t) \ge -\gamma h(x_t)
    \end{equation}
    for $0 < \gamma \le 1$, and $\Delta h(x_t, u_t) = h(x_{t+1}) - h(x_t)= h(f(x_t, u_t)) - h(x_t)$. 
    Ensuring that Eq.~\eqref{eq:cbf_cond} holds, we get that $h(x_{t+1}) \ge (1-\gamma)h(x_t)$ and the lower bound of the DCBF decreases exponentially with the decay rate $\gamma$~\cite{Sreenath_Agrawal_2017_discreteCBF}.
    One can tune the effective strength, i.e. the rate of exponential decay of the DCBF, by varying $\gamma$ between $(0,1]$. 
    
    Given a valid DCBF $h (x)$~\cite{Ames_Sreenath_Egerstedt_Tabuada_2019_CBFTheoryandApp} and imposing it as a constraint in Eq.~\eqref{eq:dcbf-def} in an optimization problem could guarantee system safety, i.e., collision-free trajectories.
    If a robotic system described by Eq.~\eqref{eq:dt-system} is safe with respect to a set $\mathcal{S} \subset \mathcal{X}$, then any trajectory starting inside $\mathcal{S}$ will remain inside $\mathcal{S}$.
    
    \begin{figure}
        \centering
        \includegraphics[width=0.47\textwidth]{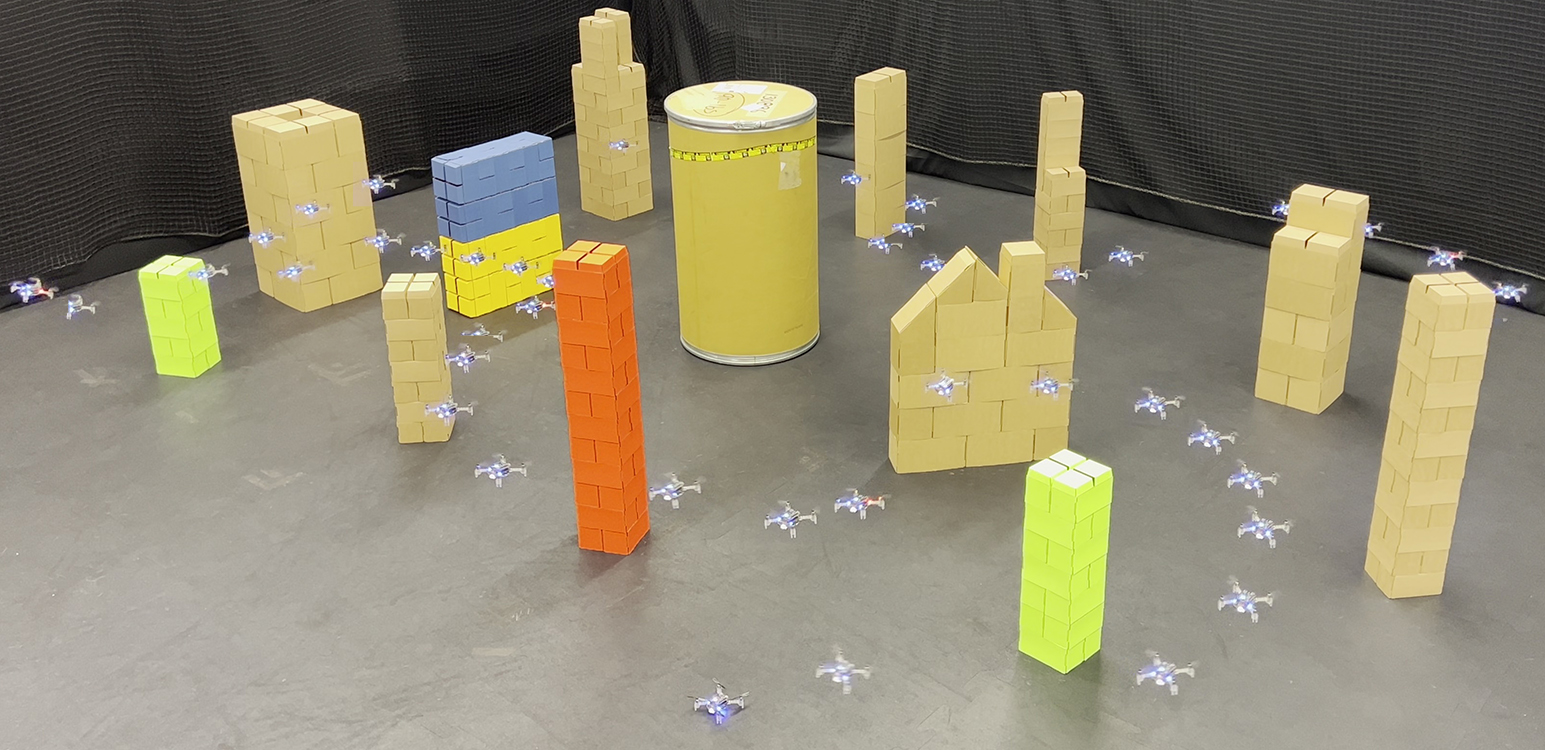}
        \caption{
        \textbf{Single-Robot Safe Exploration:} Time-lapse demonstration of a single drone robot safely navigating an exploratory plan in a cluttered environment. 
        }
        \label{fig:single_drone}
        \vspace{-5mm}
    \end{figure}

\subsection{Ergodic Exploration}\label{sec:ergodic_search}
    
    Let us consider trajectories of the state of the robot at time $t$ to be $x(t) \in \mathcal{X} \subset \mathbb{R}^n$ and the control input to the robot at time $t$ to be $u(t) \in \mathcal{U} \subset \mathbb{R}^m$. 
    In addition, let us define the robot's workspace as $\mathcal{W} \in [0,L_0]\times \ldots [0, L_{\nu-1}]$ where $L_i$ are the bounds of the workspace and $v\leq n$ is the dimensionality.
    Lastly, we define a map $g: \mathcal{X} \to \mathcal{W}$ that is continuous, differentiable, and maps the robot's state $x$ to a point in the work space (e.g., Euclidean space), that is, $g(x) = w$ and $w \in \mathcal{W}$.
    A trajectory of the robot $x(t)$ for some time horizon $t_f$ is given by solving~\eqref{eq:dt-system} for $t \in [0,t_f]$ from some initial condition $x_0$.
    
    A trajectory is said to be ergodic if its \emph{time-averaged statistics} (i.e. its spatial distribution in time) over a workspace, $\mathcal{W}$, is proportional to some measure $\phi : \mathcal{W} \to \mathbb{R}$ over the workspace\cite{Mathew_Mezic_2011_Ergodicity}\footnote{The measure $\phi$ can encode any information over the space $\mathcal{W}$, and it follows that $\int_\mathcal{W} \phi(w) dw = 1$ and $\phi(w) \neq 0$  $\forall  w \in \mathcal{W}$.}.
    For a continuous, deterministic trajectory $x(t)$, we define ergodicity as 
    \begin{equation} \label{eq:ergodicity_cont}
        \lim_{t_f\to \infty} \frac{1}{t_f}\int_{0}^{t_f} F(g(x(t))) dt = \int_\mathcal{W} \phi(w)F(w) dw
    \end{equation}
    for all Lebesgue integrable functions, $F\in \mathcal{L}^1$~\cite{Mezic_Scott_Redd_2009_DerivationErgodicityDiffScales}.
   
    We optimize trajectories $x(t)$ and control signals $u(t)$ to minimize the deviation from ergodicity in Eq.~\eqref{eq:ergodicity_cont} through the use of a Fourier transform, where we define the \emph{ergodic metric}:
    \begin{align}\label{eq:ergodic_cont}
        &\mathcal{E}(x(t), \phi) = \sum_{k\in \mathbb{N}^v} \Lambda_k \left( c_k(x(t)) - \phi_k \right)^2 \\
        &= \sum_{k\in \mathbb{N}^\nu} \Lambda_k \left( \frac{1}{t_f}\int_{t=0}^{t_f}F_k(g(x(t))) dt - \int_{\mathcal{W}} \phi(w) F_k(w)dw \right)^2 \nonumber
    \end{align}
    where $F_k(w) = \prod_{i=0}^{\nu-1} \cos(w_i k_i \pi/ L_i)/h_k$ is the cosine Fourier transform for the $k^\text{th}$ mode, $h_k$ is a normalization factor~\cite{Miller_Murphey_2016_ErgodicExploration}, and $\Lambda_k = (1 + \parallel{k}\parallel)^{-\frac{1}{2}(\nu + 1)}$ is a set of weights that penalizes lower frequency modes more.
    The subsequent trajectory optimization problem is then defined as:
    
    \begin{figure}
        \centering
        \includegraphics[width=0.5\textwidth]{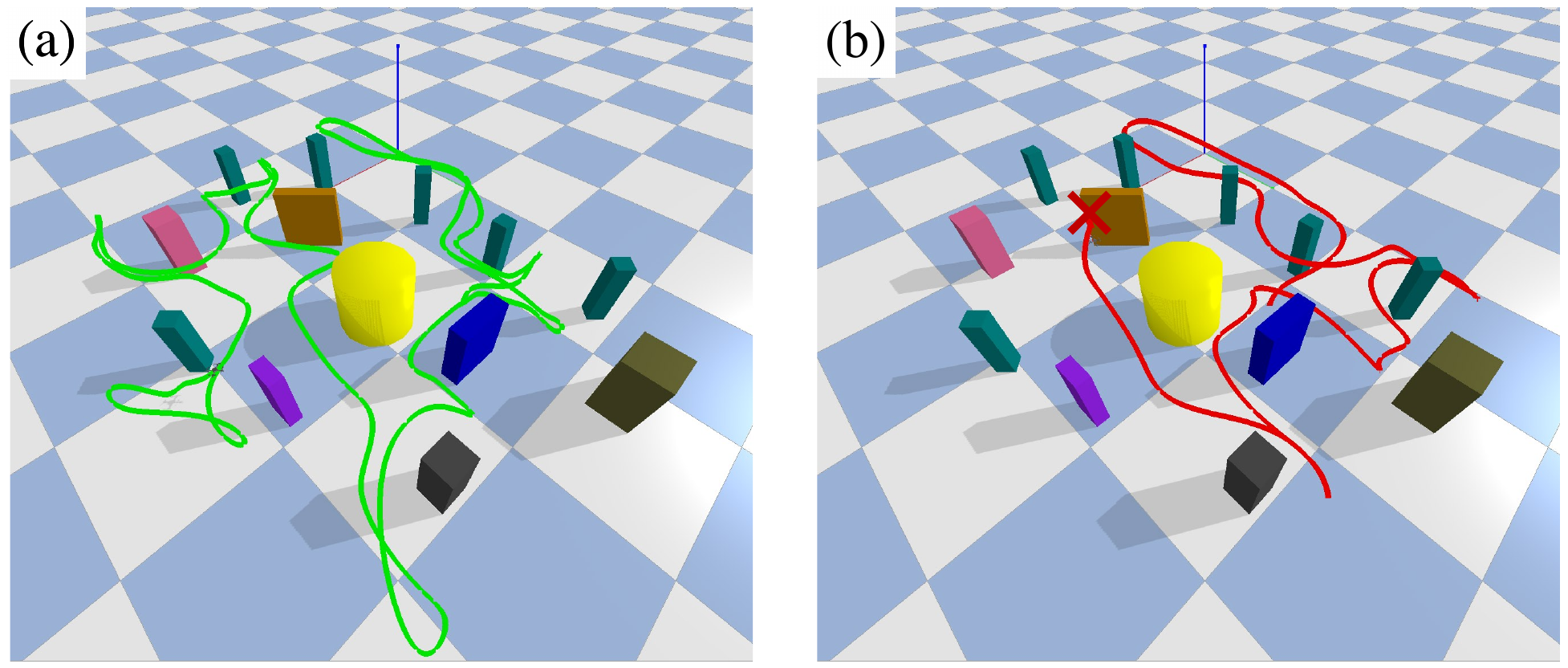}
        \begin{tabular}{*2l} \toprule
        Method & Success \% \\  \midrule
         (a) Safety-Critical ETO & 100.0 \% \\ 
         (b) ETO w/ distance constr. & 38.0\%  \\  \bottomrule
        \end{tabular}
        \caption{\textbf{Monte-Carlo Analysis:} Safety comparison between paths planned by (a) Safety Critical Ergodic Trajectory Optimization (SC-ETO) and (b) Ergodic Trajectory Optimization with distance constraints (ETO w/ distance constraints). We sample uniformly 50 pairs of random initial and final target positions. SC-ETO generates collision-free trajectories executed via a drone with a $100\%$ success rate and while ETO with only inequality constraints $h(x)$ (without the CBF condition in Eq.~\eqref{eq:dcbf-def}) generates collision free trajectories that can be executed via a drone with a 38\% success rate. }
        \label{fig:robustness}
        \vspace{-5mm}
    \end{figure}
    
    \vspace{2mm}
    \noindent
    \underline{Ergodic Trajectory Optimization:}
    \begin{subequations}\label{eq:ergodic-opt}
        \begin{align}
            &\min_{x(t),u(t)} \mathcal{E}(x(t), \phi) + \int_{0}^{t_f} u(t)^\top R u(t)dt \\ 
            &\text{s.t. } \quad
                \begin{cases}
                    \dot{x} = f(x, u),         x \in \mathcal{X}, u \in \mathcal{U} \\
                    x_{t_0} = \bar{x}_0, x_{t_f} = \bar{x}_f, g(x) \in \mathcal{W}\\
                \end{cases}
        \end{align}
    \end{subequations}
    where $R$ is a positive semi-definite matrix that penalizes control effort, and $\bar{x}_0$ and $\bar{x}_f$ are initial and final conditions. 


    \begin{figure*}
        \centering
        \includegraphics[width=\textwidth]{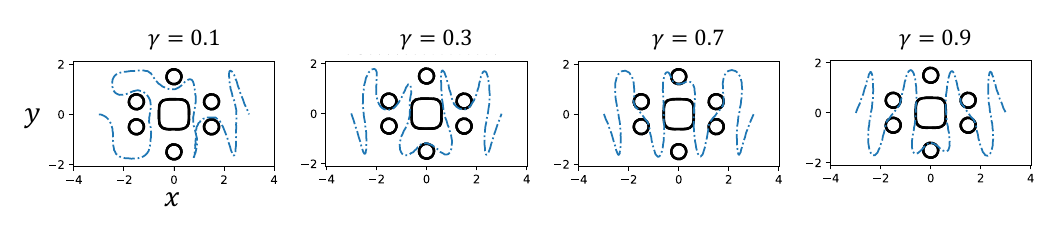}
        \caption{\textbf{Optimized trajectories for various $\gamma$ values:} 
        Each trajectory represented (blue dashed line) was optimized for the same initial and final conditions in the same exploration space. As $\gamma$ increases, the trajectories become more ergodic (see Fig.~\ref{fig:ablation2}) and follow closely to the objects. The lower the value of $\gamma$ the more cautious the trajectories become. The relationship between ergodicity and $\gamma$ is shown in Fig.~\ref{fig:ablation2}. }
        \label{fig:ablation1}
    \end{figure*}
    
    \begin{figure}
        \centering
        \includegraphics[width=0.45\textwidth]{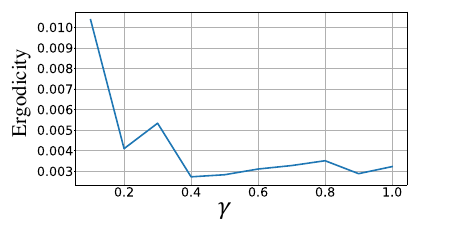}
        \caption{\textbf{Ergodicity value for various $\gamma$:}
        We evaluated ten different values of $\gamma$, ranging from 0-1. Depicted above is a trend that shows that ergodicity decreases as $\gamma$ increases (e.g. there is more coverage of the exploration space for larger values of gamma). }
        \label{fig:ablation2}
    \end{figure}

\section{Safety-Critical Ergodic Exploration} \label{sec:methods}
    Integrating control barrier functions with ergodic exploration requires that we establish the optimization in Eq.~\eqref{eq:ergodic-opt} in discrete time due to the intractability of dealing with continuous, infinite trajectories. 
    We begin by discretizing the state trajectory $\mathbf{x} =[x_0, x_1, \ldots, x_{T-1}]$, where $x_t$ is obtained through Eq.~\eqref{eq:discrete_time_f}, and the control $\mathbf{u} = [u_0, u_1, \ldots, u_{T-1}]$.
    The definition of ergodicity in Eq.~\eqref{eq:ergodicity_cont} can then be redefined for a discrete-time trajectory as:
        \begin{equation} \label{eq:ergodicity_disc}
        \lim_{T\to \infty} \frac{1}{T}\sum_{t=0}^{T-1} F(g(x_t)) = \int_\mathcal{W} \phi(w)F(w) dw
    \end{equation}
    for discrete time horizon $T$.
    With the time-averaged statistics now defined over a discrete sum, we use the same Fourier transform as before and obtain the following ergodic metric:
    \begin{align}\label{eq:ergodic_disc}
        &\mathcal{E}(\mathbf{x}, \phi) = \sum_{k\in \mathbb{N}^v} \Lambda_k \left( c_k(\mathbf{x}) - \phi_k \right)^2 \\
        &= \sum_{k\in \mathbb{N}^v} \Lambda_k \left( \frac{1}{T}\sum_{t=0}^{T-1}F_k(g(x_{t})) - \int_{\mathcal{W}} \phi(w) F_k(w)dw \right)^2 \nonumber
    \end{align}

    The discrete-time variation of ergodic trajectory optimization in Eq.~\eqref{eq:ergodic-opt} is now defined as:
    
    \vspace{2mm}
    \noindent
    \underline{Discrete-time Ergodic Trajectory Optimization (ETO)}
    \begin{subequations}
        \begin{align}\label{eq:disc_ergodic-opt}
            &\min_{\mathbf{x},\mathbf{u}} \mathcal{E}(\mathbf{x}, \phi) + \sum_{0}^{T-1} u_t^\top R u_t dt \\ 
            &\text{s.t. } \quad
                \begin{cases}
                    x_{t+1} = f(x_t, u_t), x_t \in \mathcal{X}, u_t \in \mathcal{U} \\
                    x_0 = \bar{x}_0, x_{T-1} = \bar{x}_f, g(x) \in \mathcal{W}\\
                \end{cases}
        \end{align}
    \end{subequations}

    Using the discrete-time ETO we integrate the discrete-time CBF condition in Eq.~\eqref{eq:dcbf-def} into the optimization problem. 
    Letting $h(x)$ be a valid barrier function with a defined $\gamma$, we derive the following safety-critical ETO problem statement:  
    
    \noindent
    \underline{Safety-Critical Ergodic Trajectory  Optimization (SC-ETO)}
    \begin{subequations}\label{eq:safe-ergodic-opt}
        \begin{align}
            &\min_{\mathbf{x},\mathbf{u}} \mathcal{E}(\mathbf{x}, \phi) + \sum_{0}^{T-1} u_t^\top R u_t dt \\ 
            &\text{s.t. } \quad
                \begin{cases}
                    x_{t+1} = f(x_t, u_t), x_t \in \mathcal{X}, u_t \in \mathcal{U} \\
                    x_0 = \bar{x}_0, x_{T-1} = \bar{x}_f, g(x) \in \mathcal{W}\\
                    \Delta h(x_t, u_t) \ge -\gamma h(x_t)\\
                \end{cases}
        \end{align}
    \end{subequations}
    where the DCBF is introduced into the problem as an inequality constraint.
    Solutions that satisfy Eq.~\eqref{eq:safe-ergodic-opt} result in optimized trajectories that are guaranteed to be safe and are ergodic with respect to a desired measure $\phi$ over a work space $\mathcal{W}$.

    It is possible to consider safe navigation with multiple robots in the formulation described in~\eqref{eq:safe-ergodic-opt}. 
    For each pair-wise robot, we introduce an DCBF constraint into the SC-ETO~\eqref{eq:safe-ergodic-opt} problem:
    \begin{equation}
        \Delta h(x^i_t, u^i_t, x^j_t, u^j_t) \ge -\gamma h(x^i_t, x^j_t) \, \forall i,j \in \mathcal{G}
    \end{equation}
    where $\mathcal{G}$ a fully connected graph with robot nodes $i,j$, and $\Delta h(x^i_t, u^i_t, x^j_t, u^j_t) = h(x^i_{t+1}, x^j_{t+1}) - h(x^i_t, x^j_t)$.
    Here, $h(x^i_t, x^j_t)$ is a barrier function that computes the safe and unsafe distance for two robots.
    This constraint must be satisfied along the trajectories of each robot which introduces $N(N-1)/2 \times T$ constraints (where $N$ is the number of robots and $T$ is the time horizon). 
    Assuming a homogeneous set of robots, we stack the individual states and consider the state of the system as $x = \{x^0, x^1, \ldots x^{N-1} \} \in \mathcal{X}^N \subset \mathbb{R}^{n \times N}$. 
    The stacked control input is then $u =  \{u^0, u^1, \ldots u^{N-1} \} \in \mathcal{U}^N \subset \mathbb{R}^{m \times N}$.
    We then formulate the following joint trajectory optimization problem:
    
    \vspace{2mm}
    \noindent
    \underline{Multi-Robot SC-ETO}
    \begin{subequations}\label{eq:mult-safe-ergodic-opt}
        \begin{align}
            &\min_{\mathbf{x},\mathbf{u}} \mathcal{E}(\mathbf{x}, \phi) + \sum_{0}^{T-1} u_t^\top R u_t dt \\ 
            &\text{s.t. } \hspace{-2mm}
                \begin{cases}
                    x^{i}_{t+1} = f(x^i_t, u^i_t), x^i_t \in \mathcal{X}, u^i_t \in \mathcal{U} \forall i \in \mathcal{G} \\
                    x^i_0 = \bar{x}^i_0, x^i_{T-1} = \bar{x}^i_f, g(x) \in \mathcal{W} \\
                    \Delta h(x_t, u_t) \ge -\gamma h(x_t)  \\
                    \Delta h(x^i_t, u^i_t, x^j_t, u^j_t) \ge -\gamma h(x^i_t, x^j_t) \forall (i,j) \in \mathcal{G} 
                \end{cases}
        \end{align}
    \end{subequations}

    In this paper, we assume we have a complete and connected graph $\mathcal{G}$ and overload the notation for the barrier function $h$ for ease of notation.  

    \begin{figure*}
        \centering
        \includegraphics[width=\textwidth]{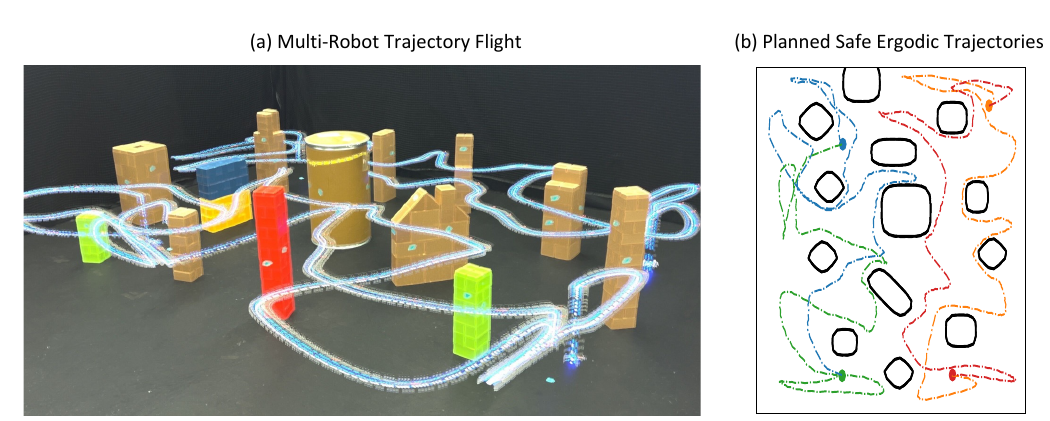}
        \caption{\textbf{Multi-Robot Safe Ergodic Exploration:} (a) Time-lapse of a multi-robot ergodic search of a cluttered exploration space and (b) the corresponding planned ergodic trajectories for each robot. 
        The four robots' trajectories are optimized jointly in a single SC-ETO (Eq.~\eqref{eq:safe-ergodic-opt}). Two pairs of drones start at opposite corners from each other and are tasked to navigate the cluttered environment, ending at the opposite corner from where they started. Inter-robot CBFs are used to avoid colliding while exploring. 
        Please view the attached multimedia to view a demonstration of this example.}
        \label{fig:multi-robot-exploration}
    \end{figure*}
    
\section{Results} \label{sec:results}

    We demonstrate the effectiveness of our approach for safety-critical ergodic exploration using a drone robotic system in a cluttered environment through two means of validation: 1) simulated results, and 2) empirical evaluations.
    For both simulated and experimental results, we assume full knowledge of the obstacle locations and shapes in the environment (as illustrated in Fig.~\ref{fig:obs_loc}) and define a uniform measure of information $\phi$ over the exploration space prior to trajectory optimization.
    We use a single integrator dynamics model when carrying out the trajectory optimization outlined in Eq. (11) with control constraints
    The drone system is simulated with a proportional, integral, derivative (PID) controller, converting planned trajectories into low-level motor commands which mimics the real system. 
    
    For all obstacles and drones, we define $h(x)$ as a signed distance function:
    \begin{equation}
        h(x) = \Big\Vert \frac{Rx-\bar{x}}{\ell+b} \Big\Vert_p - r
    \end{equation}
    where $\ell\in \mathbb{R}^n$ is a scaling factor, $\bar{x}$ is the center of the obstacle, $r$ is a radius term, $b$ is a buffer, and $p\ge2$ defines the shape of the norm (e.g., $p=4$ is more square-like), and $R$ transforms points from the world frame to the local barrier frame.
    These parameters vary according to the obstacle's dimension and shape, and are kept the same for both simulation and experiment. 
    We find the closest signed distance function that approximates the shape of each object.
    One can also mix signed distance functions to get arbitrary shapes (e.g., concave shapes) using combinations of min and max functions~\cite{sigg2003signed}.
    We use a time horizon of $T=200$ steps with a $\Delta t = 0.1$ which results in a total of $t_f=20s$ prediction horizon when solving for trajectories. 

    \subsection{Single-Drone Experimental Results}
        Our experiments are conducted using a Crazyflie 2.0 drone. We gather global position data using two IR base stations and communicate target positions with the drone via radio communication. The drone is tasked to safely explore the space defined in Fig.~\ref{fig:obs_loc}. 
        
        A time-lapse of a single-drone flight is shown in Fig.~\ref{fig:single_drone}, demonstrating that our approach generates safe, ergodic exploratory trajectories that can be run on a real-robotic system in a cluttered environment.
        SC-ETO is carried out using predefined initial and final target positions. CBFs are built around each obstacle to ensure the drone navigates safely around the environment.
        The drone tracks the optimized trajectory using an internal controller through the cluttered exploration space at a rate of 10Hz. 
        We find that the added physical constraints to the optimization problem resulted in close tracking performance on the real system.
        

    \subsection{Simulated Monte-Carlo Analysis}\label{subsec: simulated_robsust}
        To investigate the robustness of our SC-ETO method, we uniformly sample 50 randomly generated initial and final target positions in a simulated environment. 
        We simulated the drone dynamics using the pybullet gym-pybullet-drones environment~\cite{panerati2021learning} and inspect collisions during the execution of the planned paths by the drone. 
        
        We compare our SC-ETO method against ETO without the CBF constraint inequality in Eq.~\eqref{eq:cbf_cond}, which is replaced with the obstacle distance function $h(x)$. 
        Our results are presented in the table in Fig.~\ref{fig:robustness}. 
        We find that our method generates safe ergodic trajectories with a 100 percent success rate (no collisions).
        In contrast, while ETO generates collision-free trajectories using $h(x)\ge0$, the resulting trajectories are considered unsafe according to Eq.~\eqref{eq:cbf_cond}. 
        As a consequence, a drone tracking this trajectory results in unsafe exploration that collides with objects in the environment and only succeeds $38 \%$ of the time.

    \subsection{Ablation Study}
        We perform an ablation study on the CBF parameter $\gamma$ to analyze its effects on the efficacy of ergodic exploration. 
        We evaluate 10 different values of $\gamma$ between $0$ and $1$. 
        As we increase the value of $\gamma$, we find that the ergodic trajectories become less cautious and more ergodic as shown in Fig.~\ref{fig:ablation1} and in Fig.~\ref{fig:ablation2}.
        This was further reinforced by calculating the ergodicity values which we found to decrease (more coverage of a space) as $\gamma$ increases. 
        This effect can be attributed to the decay rate of the CBF as $\gamma$ decreases forcing the optimized trajectories to remain further away from the barrier. 
        Ultimately, this behavior shows a trade-off between safely navigating an environment and completely exploring all areas of the environment.
        The benefit of our approach is that we obtain this trade-off through a single value that can be tuned. 

    \subsection{Multi-Drone Exploration Experiment: }
        We further demonstrate our method on a multi-drone example. 
        A total of four drones are flown simultaneously and are tasked to safely explore the space defined in Fig.~\ref{fig:obs_loc} without colliding with one another. 
        Two pairs of drones are placed across the opposite sides of the cluttered environment and tasked to navigate to the opposite side of the environment while ergodically exploring. 
        Each pair-wise drone combination defines a CBF as a minimum distance function equal to the width of the drone to avoid collision. 
        As shown in Fig.~\ref{fig:multi-robot-exploration}, the drones are able to safely navigate and explore the environment without collision between themselves and obstacles. 
        Note that in this work, we do not numerically address the computational complexity of the multi-drone exploration problem, and leave this to future work.


\section{Conclusion and Future Work} \label{sec:conclusion}
    In this paper, we demonstrate safe and effective planning for exploration through the development of safety-critical ergodic trajectory optimization. 
    Simulated results show the robustness of our approach as a planner for generating safe ergodic exploratory trajectories. 
    Empirical evaluations demonstrate the effectiveness of our approach for safe single- and multi- drone exploration in a cluttered environment.
    Future work will focus on implementing these techniques for dynamic obstacles via model-predictive control (MPC) and integrate a more accurate model of the robot dynamics with an arbitrary number of robots.

\section*{Acknowledgments}\label{sec:acknowledgments}
    The authors would like to thank Yale's Center for Collaborative Arts and Media for their multi-media resources and Samuel Osborne for his assistance with video editing. 

\bibliography{references}
\bibliographystyle{IEEEtran}

\end{document}